\documentclass[review]{elsarticle}
\usepackage{amsmath}   
\usepackage{hyperref} 
\usepackage{algorithm}
\usepackage{algpseudocode}
\usepackage{bbding}
\usepackage{pifont}
\usepackage{wasysym}
\usepackage{amssymb}
\usepackage{tabularx}
\usepackage{amssymb}   
\usepackage{booktabs}   
\usepackage{multicol}  
\usepackage{multirow}  
\usepackage{diagbox}    
\usepackage{bm}  

\journal{Journal of \LaTeX\ Templates}   

\usepackage{graphicx}
\usepackage{caption}
\usepackage{subfigure} 
\usepackage{float}  
\usepackage[subfigure]{graphfig} 

\begin{document}

\begin{frontmatter}


\title{Discriminative Feature and Dictionary Learning with Part-aware Model for Vehicle Re-identification}  

\cortext[cor1]{Corresponding author: Xianping Fu}
\cortext[cor2]{Both Huibing Wang and Jinjia Peng contribute equally.}
\author[1]{Huibing Wang}
\author[1]{Jinjia Peng}
\author[1]{Guangqi Jiang}
\author[1]{Fengqiang Xu}
\author[1]{Xianping Fu\corref{cor1}}

\address[1]{Information Science and Technology College, Department of Computer Science and Technology, Dalian Maritime University, Dalian, Liaoning, 116026, China.}

\begin{abstract}    
	
With the development of smart cities, urban surveillance video analysis will play a further significant role in intelligent transportation systems. Identifying the same target vehicle in large datasets from non-overlapping cameras should be highlighted, which has grown into a hot topic in promoting intelligent transportation systems. However, vehicle re-identification (re-ID) technology is a challenging task since vehicles of the same design or manufacturer show similar appearance. To fill these gaps, we tackle this challenge by proposing Triplet Center Loss based Part-aware Model (TCPM) that leverages the discriminative features in part details of vehicles to refine the accuracy of vehicle re-identification.  TCPM  base on part discovery is that partitions the vehicle from horizontal and vertical directions to strengthen the details of the vehicle and reinforce the internal consistency of the parts. In addition, to eliminate intra-class differences in local regions of the vehicle, we propose external memory modules to emphasize the consistency of each part to learn the discriminating features, which forms a global dictionary over all categories in dataset. In TCPM, triplet-center loss is introduced to ensure each part of vehicle features extracted has intra-class consistency and inter-class separability. Experimental results show that our proposed TCPM has an enormous preference over the existing state-of-the-art methods on benchmark datasets VehicleID and VeRi-776.

\end{abstract}
\begin{keyword}   
\texttt  Part-aware \sep External Memory \sep Triplet-Center Loss \sep Vehicle Re-identification  
\end{keyword}
\end{frontmatter}
%
%

\section{Introduction}   

Vehicles play an essential character in the prevailing metropolitan traffic flow, so the technology associated with vehicle identification has been attracted increasing in the field of computer vision. Recently, the re-identification \cite{ref_article2,wu20193} task of capture target vehicles across cameras \cite{wang2018multiview} has been employed in the field of intelligent transportation, which intends to identify matched vehicles for a given query for the database \cite{ren2020learning,tang2019cityflow,fang2019road}. It is worth noting that the vehicle re-identification task can process the matching of large homologous images and exploit vehicle meaningful information, which has a significant driving effect on the development of smart surveillance systems\cite{ref_article5,ma2019fine}.

The vehicle retrieval task generally focuses on the locally discernable features of the vehicle, while the vehicle re-identification demands to obtain higher global features to enhance the vehicle re-identification task. Therefore, many re-identification(re-ID) methods \cite{wu2019cross,wu2018deep,wu2018and1} tend to extract vehicle global features that describe the appearance of the vehicle, which supports the same vehicle from various cameras to obtain the correct match. Despite the impressive achievement of deep learning to extract globe features in the re-ID community \cite{khorramshahi2019dual,he2019part,zhu2019vehicle}, learning a re-ID model that focuses on distinguishing local details of vehicles remains a challenge. Variations in local regions of vehicles are difficult to obtain by learning global features. Vehicles with similar appearance but different ID information have obvious differences in the partial region. Therefore, subtle differences in local vehicles may have discriminative features, which can stimulate the distinctions between different classes of vehicles. The premise of effectively extracting vehicle detailed features is to accurately locate and pay attention to the local region of the vehicle.

The development of deep learning \cite{wang2017effective,wu2020few,zhang2020twin} brings opportunities and challenges to the field of computer vision \cite{miao2018direct,miao2019lambda}. Some methods have been proposed in recent papers to obtain the discriminative details features of the local vehicle. The typical local feature extraction method to utilize the region of interest location and region segmentation. One well-known example of re-ID task is RAM \cite{liu2018ram}, a combination of the global and local features obtained by detailed visual cues in local regions. Besides, Zhao et al. \cite{zhao2019structural} proposed a vehicle re-ID algorithm based on the interest region, which combined the features of interest extracted from the local region with the depth features extracted by the classification model to improve the performance of the vehicle re-ID. Semantic partitioning based on regions of interest can maintain stable clues to accurately locate detailed features, but they are susceptible to interference from complex backgrounds. Until now, local region-based methods usually outperform global-based ones in re-ID retrieval\cite{wu2018cycle} accuracy. This is due to that global features does not consider the exclusive visual clues conveyed by local regions. In summary, improve the model's ability to perceive nuances, which will greatly help distinguish different vehicles and improve the accuracy of vehicle re-recognition.

Partial regions of the vehicle provide discriminative details information, which prompts us to focus on the vehicle's part features. In order to solve the above problems and improve the model's ability to perceive local details, this paper proposes a triple center loss based part-aware model (TCPM) to enhance the saliency of local part regions. Considering the rigid structure of the vehicle, we process the vehicle image into lower-level region parts to extract local detailed features of the vehicle. To obtain local region salient information in different spatial dimensions, this paper in the feature map to multiple partitions along the two dimensions of height and width. Partitioning the space could ensure that our network learns further significant detail information. In addition, considering the internal feature distribution of each part of the vehicle, we maintain external memory modules to store the local regional features of all vehicle samples separately. In particular, the triplet center loss function is adopted to extend the difference between and within vehicle classes to obtain a further robust vehicle re-identification performance. The major contributions of our work are as follows:

\begin{itemize} 

\item By leveraging external memory modules to store the features of each part region as a global dictionary, the feature distribution of the entire dataset is modeled and the global central feature vector is obtained. In addition, the triplet central loss uses the global central feature vector as the anchor point to measure the feature vector of the storage module to obtain the strongest re-ID performance. 

\item We propose a novel triplet center loss based part-aware model to extract local detail features from two space dimensions of vehicle images separately, which strengthen the internal consistency of the vehicle image and extract discriminative features by partitioning strategy. Experiments demonstrate that our method obtains competitive performance on benchmark datasets.

\end{itemize}

The rest of this article is organized as follows. In second 2, We summarize and discuss related work on vehicle re-ID task. Section 3 introduce the core methods. Section 4 discusses the experimental results of two vehicle re-ID data sets, and section5 conclusions our proposed method.

\section{Related work}    

\subsection{Vehicle re-ID}

With the prosperity continuous evolution of feature extractors, vehicle re-identification tasks have obtained further attention in recent years. This section reviews the existing vehicle re-identification tasks. Zapletal et al. \cite{zapletal2016vehicle} used color histograms and histograms of oriented gradients to solve the vehicle re-identification problem by a linear regressor. This is one of the earliest known researchers who have conducted research on vehicle re-identification tasks. However, the traditional hand-crafted feature extractor cannot extract discriminative features, and there is no substantial generalization ability for the robustness of large datasets and models. Furthermore, Liu et al. \cite{liu2016large} adopted a multi-feature fusion method to fuse vehicle color, texture, and deep learning semantic features to identify the vehicle. Due to the continuous development of deep learning \cite{simonyan2014very} \cite{szegedy2015going}, the extracted deep features has been made significant progress to target recognition tasks. In \cite{zhou2018aware}, the viewpoint information is utilized to construct a viewpoint-sensitive framework and transform the single-view features into global multi-view features for representation. In \cite{li2017deep}, the authors designed a unified framework, tasks such as recognition, attribute recognition, and verification are combined to extract the discriminative features of the vehicle.

In general, vehicle re-identification ordinarily adopts the whole graph to obtain global feature representations for image retrieval tasks, which will incurs to bottlenecks in accuracy. Therefore, some researchers have focused on local regions of vehicles to obtain discriminate features.Liu et al. \cite{liu2018ram} proposed a model consisting of global and locally discriminable features, called a region-aware deep model (RAM), which extracts features from multiple overlapping local regions and fuses more visual clues. In addition, Zhao et al. \cite{zhao2019structural} proposed a vehicle re-identification method based on the region of interest, which combined the ROI local features into a structural feature for vehicle re-identification tasks. Wang et al. \cite{wang2017orientation} adopted the method of keypoint location to apply the 20 key points of the image as key identification features, in which local regions are segmented and space-time constraints are modeled to improve retrieval performance. But the key points demand to incease manual annotation, which will boost a lot of extra workloads.
Peng et al. \cite{peng2019learning} proposed a multi-region model(MRM) to extract feature of local regions, which adopt spatial transformer network to locate the local region for learning discriminate feature for vehicle re-identification. Ma et al. \cite{ma2019vehicle} proposed an improved partial model to learn deep features, which refined the local features and divided the local features through a grid spatial transformer network. In summary, local regions play an essential role in vehicle re-identification tasks and have captivated the attention of many scientific researchers.


\subsection{Distance Metric Learning}

Yan Bai et al. \cite{bai2018group} proposed a sensitive triplet embedding method to handle intra-class variance and gave a mean-valued triplet loss to alleviate the negative effects of improper triplet sampling during the training. Guo et al. \cite{guo2019two} proposed an attention network to enhance the discrimination ability between different models to optimize vehicle re-identification performance. Liu et al. \cite{liu2016deep} proposed a deep relative distance learning method (DRDL) for vehicle re-identification. They used coupled cluster loss to divided images of different vehicles from images of the same vehicle by a certain distance. Wen et al. \cite{wen2016discriminative} proposed center loss to reduce the distance within each category, which updates and records the center of each category in each iteration.

\section{Triplet Center Loss based Part-aware Model}

\subsection{Overview}

\begin{figure}[tp]
\includegraphics[width=12cm]{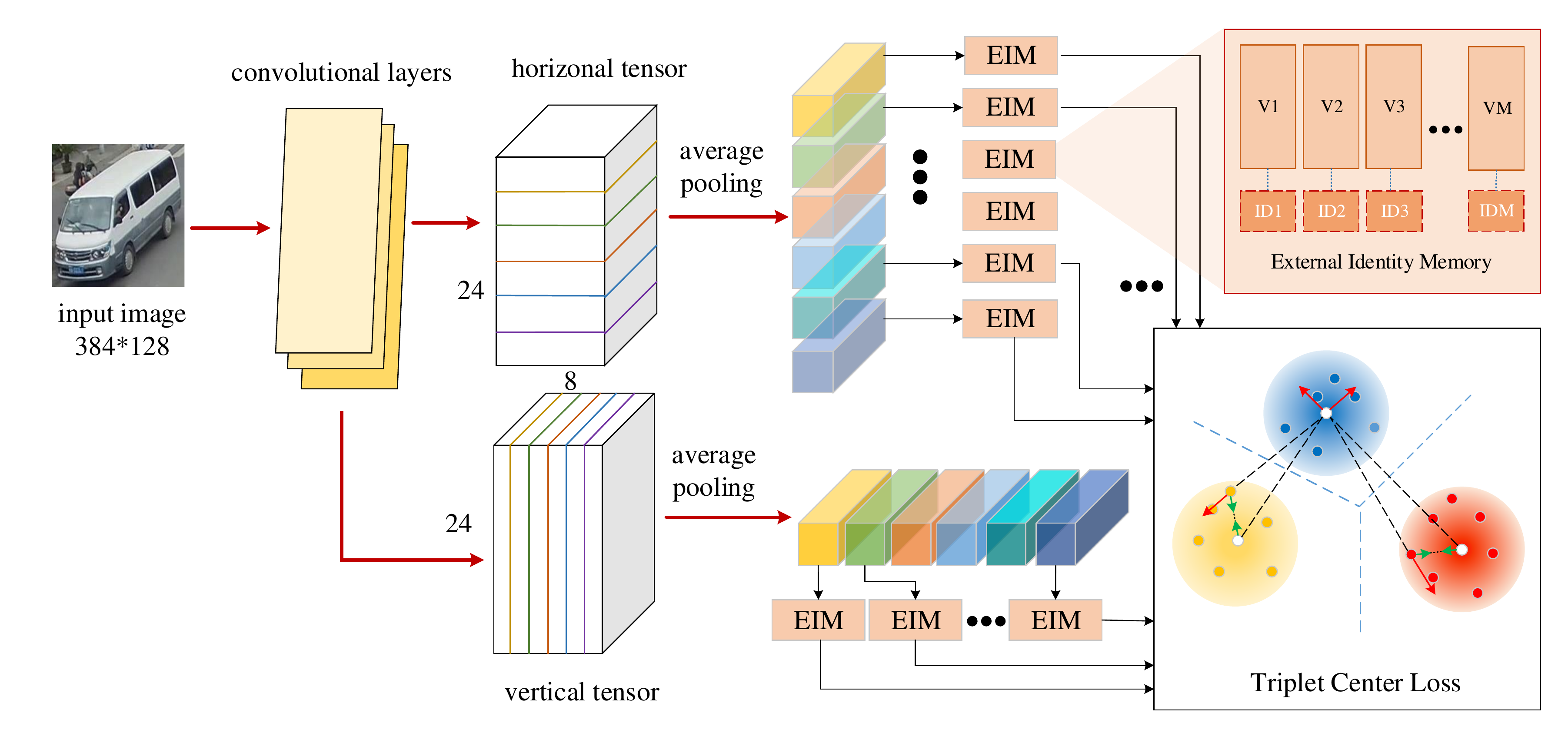}
\caption{Triplet Center Loss based Part-aware Model. The proposed model extracts feature from a deep convolutional network, which will be copied into two 3D tensors as two branches. The tensors of these two branches will be partitioned from vertical orientation and horizontal orientation into six sub-regions. Then the external memory module will be adapted to store the part features of the two branches separately. Finally, the features stored in each memory module are measured to utilize a triple-center loss and a softmax loss for metric learning to train the re-ID model.}
\label{framework}       
\end{figure}

\begin{figure*}[htbp]
	\includegraphics[width=12cm]{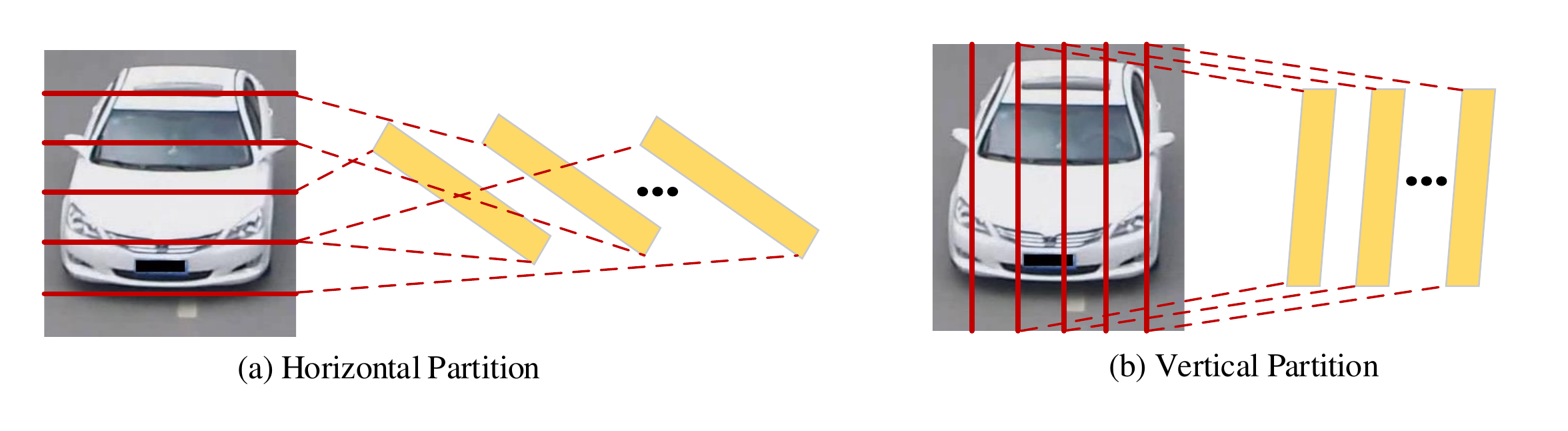}
	\caption{The illustration of horizontal partition and vertical Partition.}
	\label{partition}
\end{figure*}

In this paper, we mainly focus on leveraging the discriminative features in part details of vehicles to improve the accuracy of vehicle re-identification. Therefore, we utilize the partitioned block method to strengthen the details of the vehicle and reinforce the internal consistency of the parts. In order to combine block partition and metric learning for vehicle re-identification, we propose a Triplet Center Loss based Part-aware Model(TCPM) illustrated in Fig.\ref{framework}. In our model, we utilize competitive ResNet50 as the backbone network to learn global features from the vehicle image. The input images are first feed-forward into the four convolution blocks of the backbone network to pre-train. In order to obtain the detailed features of different spatial dimensions of the vehicle, we consider spatial partitioning the feature map from the height and width directions respectively.The region part module is shown in Fig.\ref{partition}.

Specifically, we duplicate the last layer of the backbone network and split it into two branches, which the two three-dimensional tensors are generated were represented $T$ and $T^T$, respectively. Where $T^T$ represents the transpose of $T$. Here, we name the two branches as horizontal part-aware branch and vertical part-aware branch. So feed an input vehicle image from the backbone network, the part-aware model could obtain $T$ and $T^T$, which can be partitioned from vertical orientation and horizontal orientation into six sub-regions. For the $i$-th sample, the three-dimensional part-aware tensor can be formulated as:

\begin{equation}\begin{array}{l}
\left\{ \begin{array}{l}
T\{ i\}  = [T(i,1),T(i,2),...T(i,{p_1})]\\
{T^T}\{ i\}  = [{T^T}(i,1),{T^T}(i,2),...{T^T}(i,{p_2})]
\end{array} \right.\\
s.t.  {\rm{ }}i = 1,2...M
\end{array}
\end{equation}

where $M$ is the number of samples in dataset, $p_1$ and $p_2$ represent the number of first and second branch parts, respectively. We name $T$ and $T^T$ as the horizontal and vertical tensors, respectively. This operates not only guarantees that the network can effectively learn the fine-grained local features with spatial information but also emphasize the consistency within each part of the region. Subsequently, the conventional average pooling is adopted for each partitioned sub-region to obtain the feature vector $h$ with the size of 2048 × 1. Specifically, the feature vector $h$ could be represented as follows:

\begin{equation}\begin{array}{l}
h\{ i\}  = [h(i,1),h(i,2),...h(i,b)]\\
s.t. {\rm{ }} i = 1,2,...M ;  {\rm{   }}b = {p_1} + {p_2}
\end{array}\end{equation}

where $b$ is the total number of parts. In addition, considering the internal feature distribution of vehicle data, the external memory module is exerted to store the horizontal and vertical orientation feature vectors in each memory bank. For each memory module, we store $h$ and $ID$ information of each part, which are recorded as  $V=[{v_1,v_2,...v_M}]$ and $ID=[{id_1,id_2,...id_M}]$, respectively. The feature distribution of each partial region can be modeled and the feature vector clustering center under each subregion can be obtained by loss function. Finally, adopt Triplet-Center Loss and softmax loss to the features stored in all memory modules effectively minimizes the intra-class distance of deep learning features, while maximizing the inter-class distance of features.

\subsection{ Vehicle Re-identification with External Memory}

Invariance constraints imposed by memory module on the sample recently received widespread attention, organize the features in the form of structured and addressable become the processing features of the essential steps. In our paper, from another perspective, emphasizes the consistency of each part, which we speculate the adopt of memory is essential to reduce intra-class differences in local regions of the vehicle image. We introduce an External Identity Memory into the network to store the features of the two branches of a total of 12 sub-regions and their $ID$ information, which can implement invariance constraints for each sub-region to reduce intra-class differences the vehicle part module.

External Identity Memory composed of two components including partial feature External Memory and identify memory: We maintain a partial feature External Memory for storing six distinguished sub-region features of each branch, denoted as $V=[{v_1,v_2,...v_M}]$. Where $M$ is the number of images in the vehicle dataset. Besides, identify memory is designed to preserved the label information corresponding to sub-region each feature denoted as $ID=[{id_1,id_2,...id_M}]$.
We initialize the values of feature vectors of the memory module, where the feature vectors of all subregions of each image instance are treated as an individual category. The sub-regions of the feature vector segmented by the average pooling layer can be represent is ${h\{i,b\}}=[{h(1,b)},{h(2,b)},...{h(M,b)}] $, Where $b$ represents the $b$-th partial subregions, and there are 12 subregions in our paper. During each training iteration, the feature vectors are continuously stored in the memory module, which can obtain the feature distribution of each partial region. During the backpropagation, we update the features of the training sample ${h\{i,b\}}$ in memory by:
\begin{equation}
{V_{i,b}} \leftarrow \delta {V_{i,b}} + (1 - \delta ){h\{i,b\}}
\end{equation}

Where $ {V_{i,b}}$ is the memory of sample in $b$-th subregions of $i$-th samples, which uses the hyper-parameter $\delta \in [0,1]$ to control the update rate. Therefore, the memory module is utilized to constrain the inter-class invariance of the local regions of the vehicle to improve the performance of vehicle identification.

\subsection{ Global Dictionary Learning}

Through the external memory, we construct a global dictionary for all categgories in the dataset.  The memory enables to compare the training samples of each part of the samples with the global training samples and implement the global invariance constraint by modeling the feature distribution of the whole vehicle datasets. Hence, we regard the memory as the global dictionary to capture patterns commonly shared across classes.
For a given number of categories $R$, we learn the unique memroy for each category. Based on the learned global dictionary, an image could be compactly described by the feature from the corresponding memroy. Consider a database of $N$ images with $R$ classes. Training samples $\{s_i\}_{i=1}^N$ could be represented as $\{m_i\}_{i=1}^N$, where $m_i$ is the feature from memory structure. The features would be utilized as the discriminative features to calculate the probability of similar vehicles.

\subsection{Triplet-Center Loss Function}
In TCPM, we devise the triplet-center loss function and softmax loss function to learn robust and discriminative features. Triplet-center loss was maintained in Part-aware Model primarily to ensure each part of features extracted by the network have intra-class consistency and inter-class separability. In our model, the new sample will perform similarity calculations with data samples of the entire dataset stored by the memory model instead of a batch sample. The memory model obtains the average value of each class of samples of each sub-region part through the network iterative updating as the center feature vector of triplet-center loss. Then combine softmax loss function to obtain discriminative features. The objective function can be written as：

\begin{equation}
L = \lambda {L_{tcl}} + {L_{soft\max }}
\end{equation}

Where ${L_{tcl}} $ represent the triplet-center loss function and $ {L_{soft\max }} $ sofemax loss in our model, respectively. Where $\lambda$ is a hyper-parameter to balance the weight between triplet-center loss and softmax loss.

The features representation $f \{i, b\}$ with the associated ground truth identity ${y_i}$ of the $i$-th sample in the $b$ region is obtained through the full connection layer. We assume that each block sample from the same class share a corresponding clustering center. Therefore, we can obtain the set of center feature vector $C\{i, b\}$ with label $l$, where $C\{i, b\}$ represents the $i$-th sample and the $b$-th region. For training data containing $M$ samples and $12$ subregions, the loss function of $b$-th subregion can be described as:

\begin{equation}
{L_{tcl,b}} = \sum\limits_{i = 1}^M {\max (d(f\{ i,b\} ,c\{ {l^i},b\} ) + \alpha  - \mathop {\min }\limits_{j \ne i} d(f\{ i,b\} ,c\{ j,b\} ),0)}
\end{equation}
Where $d$ is the square of the $l_2$-norm in the sample clusters, and its calculation formula is:

\begin{equation}
d(f\{ i,b\} ,c\{ {l^i},b\} ) = \frac{1}{2}\left\| {f\{ i,b\}  - c\{ {l^i},b\} } \right\|_2^2
\end{equation}
The center of our update parameters in each iteration£¬which based on each subregion batch of data.

In order to better optimize the cluster center and effectively separate samples of different classes, we combine the loss function of metric learning and classification tasks. Given the feature vector of each part $f \{i,b\}$, first calculate the cosine similarity between the features of $f\{i, b\}$ and the features stored in the external memory. Then, the softmax function is applied to calculate the prediction probability of $x_{i}$ belonging to class $i$. It can be described as:

\begin{equation}
{L_{soft\max }} =  - \log \frac{{\exp ({V_{i,b}}^Tf\{ i,b\} /\beta )}}{{\sum\limits_{j = 1}^N {\exp ({V_{j,b}}^Tf\{ i,b\} /\beta )} }}
\end{equation}

Where ${{V_{i,b}}}$ is the memory of image $x_{i}$ in $i$-th slot of $b$ subregion. $N$ is the number of classes of the training set. Where $\beta \in (0, 1]$ is a hyperparameter that balances the scale of the distribution.

\section{Experiments}

In this section, we evaluate our proposed method for vehicle Re-ID using the Cumulative Match Characteristic (CMC) curve and mean Average Precision (mAP) \cite{lin2019improving} widely adopted in vehicle ReID. Besides comparing with state-of-the-art vehicle ReID methods, a series of detailed studies are conducted to explore the effectiveness of proposed method. All the experiments are conducted on two vehicle ReID datasets: VeRi-776 \cite{liu2017provid} and VehicleID \cite{liu2016deep}.

\subsection{Datasets}

\begin{itemize}
	\item VeRi-776  \cite{liu2017provid}. VeRi-776 is a large-scale urban surveillance vehicle dataset captured from real-world scenes for re-ID, which contains more than 50,000 images of 776 vehicles with identity annotations, image timestamps, camera geo-locations, vehicle color and type information. Each vehicle is from 2-18 cameras with various viewpoints, illuminations and occlusions. In this dataset, 37,781 images of 576 vehicles are split as a train set and 11,579 images of 200 vehicles are employed as a test set. A subset of 1,678 images in the test set generates the query set.
	
	\item VehicleID \cite{liu2016deep}. It is a widely-used vehicle reID dataset, which contains 26267 vehicles and 221763 images in total. The training set contains 110,178 images of 13,134 vehicles. From the original testing data, four subsets are extracted, which contain 800, 1,600, 2,400 and 3,200 vehicles, and are searched in different scales. During the phrase of testing, an image is randomly selected from one identity to obtain a gallery set with 800 images, and then the remaining images are all employed as probe images. Two other test sets are processed in the same way.
\end{itemize}

\subsection{Evaluation metrics}

The mAP \cite{lin2019improving} and CMC curve are employed to evaluate the performance of our proposed TCPM in this paper. For each query image in a subset of test images, it would be tested with other test images, the average precision for each query $q$ is calculated by
\begin{equation}
AP(q)=\frac{\sum_{k=1}^nP(k)\times{rel(k)}}{N_{gt}}
\end{equation}

Where $P(k)$ denotes the precision at the $k_{th}$ position of the results. The $rel(k)$ is an indicator function equal to 1 if the $k_{th}$ result is correctly matched or zero otherwise. $n$ is the number of tests, and $N_{gt}$ is the ground truths. After experimenting for each query image, the $mAP$ will be calculated as follows:
\begin{equation}
mAP=\frac{\sum_{q=1}^QAP(q)}{Q}
\end{equation}

where $Q$ is the number of all queries. In this paper, the vehicle images with the same ID and camera number are considered to be junk images in our evaluation of results.

\subsection{Implementation Details}
The proposed method is trained in the Pytorch \cite{paszke2017automatic}. Besides that, the ablation experiments are all conducted in the PyTorch. Stochastic gradient descent (SGD) \cite{bottou2010large} is employed to update the parameters of the network with a momentum of $\mu=0.9$ during the training procedure on both VehicleID and VeRi-776. Due to the limit of the memory, besides the proposed network, the batch size of other experiments is set to 32. 16 is set when training the proposed network. The learning rate of each epoch is various.  the first 40 epochs is set to 0.05 while the last 20 is 0.005.

\subsection{Comparison with the state-of-the-art methods}

\subsubsection{Evaluation on VeRi-776} The results of the proposed method is compared with state-of-the-art methods on VeRi-776 dataset in Tables \ref{tab1} \ref{tab2}, which includes: (1) LOMO \cite{liao2015person}; (2) DGD \cite{xiao2016learning}; (3) GoogLeNet \cite{yang2015large} (4) PROVID \cite{liu2017provid}; (5) PathLSTM \cite{shen2017learning}; (6) OIFE+ST \cite{wang2017orientation}; (7) GSTE \cite{bai2018group}; (8) VAMI \cite{zhou2018aware}; (9) VAMI+ST \cite{zhou2018aware}; (10) RAM \cite{liu2018ram}; (11) AAVER \cite{khorramshahi2019dual}; (12) Part-Regular \cite{he2019part}. The performance comparisons on VeRi-776 dataset have been listed in the Table \ref{tab1}. From Table \ref{tab1}, it is obvious that our proposed method achieves the best performance with rank-1 = 93.98\%, rank-5= 97.13\%, mAP = 74.59\% on VeRi-776. Especially, some of methods exploit utilizing other semantic information to train the reID model, such as RAM\cite{liu2018ram}, AAVER \cite{khorramshahi2019dual}. Compared with these methods, our method focus on learning details features only from the identity information, which also have significant improvements on VeRi-776.

\begin{table}[htbp]
	\renewcommand{\arraystretch}{1.3}
	\centering
	\scriptsize
	\caption{Experimental results on VeRi-776. The mAP ($\%$) and cumulative matching scores ($\%$) at rank 1, 5 are listed.}\label{tab1}
	\begin{tabular}{p{4.3cm}|p{1.0cm}|p{1.0cm}|p{1.0cm}}
		\hline
		Method & mAP & Rank1 & Rank5\\
		\hline
		\hline
		LOMO \cite{liao2015person} &  9.64 & 25.33 & 46.48\\
		DGD \cite{xiao2016learning} &  17.92 & 50.70 & 67.52\\
		GoogLeNet \cite{yang2015large} &  17.81 & 52.12 & 66.79\\
		PROVID \cite{liu2017provid} &  53.42 & 81.56 & 95.11\\
		PathLSTM \cite{shen2017learning} & 58.27 & 83.49 & 90.04\\
		OIFE+ST \cite{wang2017orientation} & 51.42 & 68.30 & 89.70\\
		GSTE \cite{bai2018group} &59.47 &96.24 &98.97 \\
		VAMI \cite{zhou2018aware} & 50.13 & 77.03 & 90.82\\
		VAMI+ST \cite{zhou2018aware} & 61.32 & 85.92 & 91.84\\
		RAM \cite{liu2018ram} &61.5 &88.6 &94.0 \\
		AAVER \cite{khorramshahi2019dual} &66.4 &90.2 &94.3 \\
		Part-Regular \cite{he2019part}  &74.3 &94.3 &98.7 \\
		\hline
		\hline
		TCPM &  74.59 & 93.98 & 97.13\\
		\hline
	\end{tabular}
\end{table}

Compared with those reID model trained by hand-crafted features, such as LOMO \cite{liao2015person} and DGD \cite{xiao2016learning}, our method have 64.95$\%$ and 56.67$\%$ improvements in mAP, respectively. This verifies that the features obtained from deep model are more robust than the hand-crafted feature that are severely affected by the complicated environment. Some method consider the spatio-temporal information to better constrain the re-ID model, which is also provided in the VeRi-776, such as PROVID \cite{liu2017provid}, PathLSTM \cite{shen2017learning}, OIFE+ST \cite{wang2017orientation} and VAMI+ST \cite{zhou2018aware}. Compared with those methods, our methods also achieve better performance without any other information besides the identity information, which verifies the effectiveness of extracting the discriminative feature by the proposed method. At last, compared with the RAM \cite{liu2018ram} that has similar structure with our method, there are 13.09$\%$ and 5.38$\%$ in mAP and rank-1 gains on VeRi-776. It strongly proves that besides the proposed network could achieve more details, the triplet center loss also could constrain the relationship among the inter-class and intra-class to better train the re-ID model with the proposed network. It is worth mentioning that RAM \cite{liu2018ram} also concentrates on the attribute training for optimizing the feature representation to improve the reID performance, while our method pays more attention to improving the accuracy from exploiting more details.

\begin{table}[htbp]
	\renewcommand{\arraystretch}{1.2}
	\centering
	\scriptsize
	\caption{Experimental results on VehicleID. The mAP ($\%$) and cumulative matching scores ($\%$) at Rank 1, 5 are listed.}
	\begin{tabular}{p{2.1cm}|p{0.7cm}|p{0.8cm}|p{0.8cm}|p{0.7cm}|p{0.8cm}|p{0.8cm}|p{0.7cm}|p{0.8cm}|p{0.8cm}}
		\hline
		\multirow{2}*{Method} & \multicolumn{3}{c|}{Test size = 800} & \multicolumn{3}{c|}{Test size = 1600}& \multicolumn{3}{c}{Test size = 2400} \\ 
		\cline{2-10} &  mAP & Rank1 & Rank5 & mAP & Rank1 & Rank5 & mAP & Rank1 & Rank5 \\
		\hline
		\hline
		BOW-SIFT \cite{liu2016deep2}   & -  & 2.81  & 4.23 &- & 3.11 & 5.22 & -	& 2.11	& 3.76\\
		LOMO \cite{liao2015person}  & -  & 19.76  & 32.14 &- & 18.95 & 29.46 & -	& 15.26	& 25.63\\
		DGD \cite{xiao2016learning}  & -  & 44.80  & 66.28 &- & 40.25 & 65.31 & -	& 37.33	& 57.82\\
		Mixed DC \cite{liu2016deep}   & -	&49.0	&73.5	&-	&42.8	&66.8	&-	&38.2&	61.6\\
		FACT  \cite{liu2017provid}  & -	&49.53	&67.96	&-	&44.63	&64.19	&-	&39.91&	60.49\\
		NuFACT \cite{liu2017provid}  & -	&48.90	&69.51	&-	&43.64	&65.34	&-	&38.63&	60.72\\
		OIFE \cite{wang2017orientation}   &-	&-&	-	&-	&-	&-	&-	&67.0	&82.9\\
		VAMI \cite{zhou2018aware} &-	&63.12&	83.25	&-	&52.87	&75.12	&-	&47.34	&70.29\\
		TAMR \cite{guo2019two}    &67.64	&66.02 &79.71	&63.69	&62.90	&76.80	&60.97	&59.69	&73.87\\
		AAVER \cite{khorramshahi2019dual} &- &74.69 &93.82 &- &68.62 &89.95 &- &63.54 &85.64 \\
		HV-EALN \cite{lou2019embedding} &77.50 &75.11 &88.09 &74.20 &71.78 &83.94 &71.00 &69.30 &81.42 \\
		\hline
		\hline
		TCPM &85.13 &81.96	&96.38 &82.12 &78.82 &94.29	&78.05 &74.58 &90.71\\
		\hline
	\end{tabular}
\label{tab2}
\end{table}

\subsubsection{Comparison on VehicleID} There are 11 methods are compared with our proposed method, which are (1) BOW-SIFT \cite{liu2016deep2}, (2) LOMO \cite{liao2015person}, (3) DGD \cite{xiao2016learning}, (4) Mixed DC \cite{liu2016deep}, (5) FACT \cite{liu2017provid}, (6) NuFACT \cite{liu2017provid}, (7) OIFE \cite{wang2017orientation}, (8) VAMI \cite{zhou2018aware}, (9) TAMR \cite{guo2019two}, (10) AAVER \cite{khorramshahi2019dual}, (11) HV-EALN \cite{lou2019embedding}. Table \ref{tab2} illustrates the rank-1, rank-5 and mAP of our method and other comparison methods on VehicleID. Different VeRi-776, there is no spatio-temporal labels in VehicleID. Hence, there are no methods that consider the spatio-temporal information. All compared methods utilize the appearance information only from vehicle images. The proposed method outperforms all deep learning based methods under comparison on the test sets with different sizes on VehicleID, which obtains $81.96\%$, $78.82\%$, $74.58\%$ in rank-1, respectively.  And this also shows that our proposed method could generate more distinct features for different vehicle reID datasets.

The proposed methods shows a larger accuracy improvement when compared with traditional methods, such as LOMO \cite{liao2015person} and DGD \cite{xiao2016learning}. There are $79.15\%$ and $62.2\%$ gains in rank-1 on the test set 800, respectively. The similar improvements also occur on other test sets. Compared with those methods that learn multi-view features, the proposed also show satisfactory performance. For instance, compared with VAMI \cite{zhou2018aware}, our method has a gain of $18.84$, $25.85$, $27.24$ in terms of rank-1 accuracy on different test sets.

Moreover, for those methods which also exploit the details by attention network or detection network, such as TAMR \cite{guo2019two} and HV-EALN \cite{lou2019embedding}, our method also achieve competitive results. Compared with these methods, our method is more easer to implement and doesn't need any other annotations. Especially, some annotations are not labeled in some existing network.

\subsection{Ablation Studies}
To validate the necessity of the proposed method, some ablation experiments are conducted. The comparison results on VeRi-776 and VehicleID are presented in Table \ref{tab3} and Table \ref{tab4}. ``Baseline'' means the reID model is trained by the resnet50 with triplet loss. ``OneBranch'' represents that there is only one branch with triplet loss in the training network. ``OneBranch+Mem'' is the same backbone network with ``OneBranch'' and adds the memory structure into the network. It is also trained with triplet loss. The only difference between ``OneBranch+Mem+TC'' and ``OneBranch+Mem'' is the training loss. ``TC'' means the triplet center loss. ``TwoBranch+Mem+TC'' is the proposed network in our paper.

\begin{table}[htbp]
	\renewcommand{\arraystretch}{1.3}
	\centering
	\scriptsize
	\caption{Performance of features fusion on VeRi-776. The mAP ($\%$) and cumulative matching scores ($\%$) at Rank 1, 5 are listed.}
	\begin{tabular}{p{3.5cm}|p{1.0cm}|p{1.0cm}|p{1.0cm}}
		\hline
		Descriptor & mAP & Rank1 & Rank5\\
		\hline
		\hline
		Baseline &  62.71 & 91.53 & 95.53\\
		OneBranch &  71.64 & 92.80 & 96.43\\
		OneBranch+Mem &  72.27 & 91.97 & 95.90\\
		OneBranch+Mem+TC &  73.32 & 92.86 & 96.43\\
		TwoBranch+Mem+TC &  74.59 & 93.98 & 97.13\\
		\hline
	\end{tabular}
\label{tab3}
\end{table}

\begin{table}[htbp]
	\renewcommand{\arraystretch}{1.3}
	\centering
	\scriptsize
	\caption{Experimental results on VehicleID. The mAP ($\%$) and cumulative matching scores ($\%$) at Rank 1, 5 are listed.}
	\begin{tabular}{p{3cm}|p{0.7cm}|p{0.7cm}|p{0.7cm}|p{0.7cm}|p{0.7cm}|p{0.7cm}|p{0.7cm}|p{0.7cm}|p{0.7cm}}
		\hline
		\multirow{2}*{Method} & \multicolumn{3}{c|}{Test size = 800} & \multicolumn{3}{c|}{ Test size = 1600}& \multicolumn{3}{c}{ Test size = 2400} \\ \cline{2-10} &  mAP & Rank1 & Rank5 & mAP & Rank1	& Rank5 & mAP & Rank1 & Rank5 \\
		\hline
		\hline
		Baseline   & 80.07  & 77.27  & 90.42 &76.50 & 73.71 & 85.99 & 73.68	& 70.98	& 82.87\\
		OneBranch  & 82.14 & 79.69  & 94.03 &78.97 & 76.63 & 91.58 & 74.19	& 71.48	& 88.60\\
		OneBranch+Mem  & 83.42  & 80.16  & 96.03 &80.24 & 76.94 & 93.86 & 76.15	& 72.73	& 89.50\\
		OneBranch+Mem+TC   & 84.62	&81.50	&96.27	&81.53	&78.22	&93.95	&77.16	&73.74&	89.54\\
		TwoBranch+Mem+TC  &85.13 &81.96	&96.38 &82.12 &79.02 &94.89	&77.95 &73.88 &90.82\\
		\hline
	\end{tabular}
\label{tab4}
\end{table}

Firstly, compared with ``Baseline'', our proposed method ``TwoBranch+Mem+TC'' has a large improvement in mAP, rank-1 and rank-5. This is because that ``Baseline'' only consider the global features for reID task while our method exploits the details. Through the comparison of ``OneBranch'' and ``OneBranch+Mem'', the results could prove that the memory structure could balance the relationship between single images and the group images with the same identity. Besides that, to verify the effectiveness of the triplet center loss, the ``OneBranch+Mem+TC'' and ``OneBranch+Mem'' are conducted. As reported in Table \ref{tab3} and Table \ref{tab4}, it is worth noting that, both on the VeRi-776 and VehicleID, the ``OneBranch+Mem+TC'' has improvements in different test sets. For instance, there is 1.05\% and 0.89\% gains in mAP and rank-1 on VeRi-776, respectively. And for VehicleID, it also improves 1.2\%, 1.29\% and 1.01\% in mAP on different test sets.

Moreover, to verify the proposed network with different branches, the comparison of ``OneBranch+Mem+TC'' and ``TwoBranch+Mem+TC'' is designed. The only difference of them is that the ``TwoBranch+Mem+TC'' has the vertical branch, which divides the global feature into several parts in the vertical direction. On VeRi-776, compared with ``OneBranch+Mem+TC'', ``TwoBranch+Mem+TC'' has gains of 1.27\%, 1.12\% in mAP and rank-1, respectively. For VehicleID, we also observe improvements of 0.51\%, 0.59\%, 0.79\% in mAP on test set with the size of 800, 1600 and 2400. All of these show that the proposed network could learn more discriminative features for vehicle re-ID.

\subsection{Visualization of Results}

\begin{figure}[htbp]
	\includegraphics[width=12.5cm]{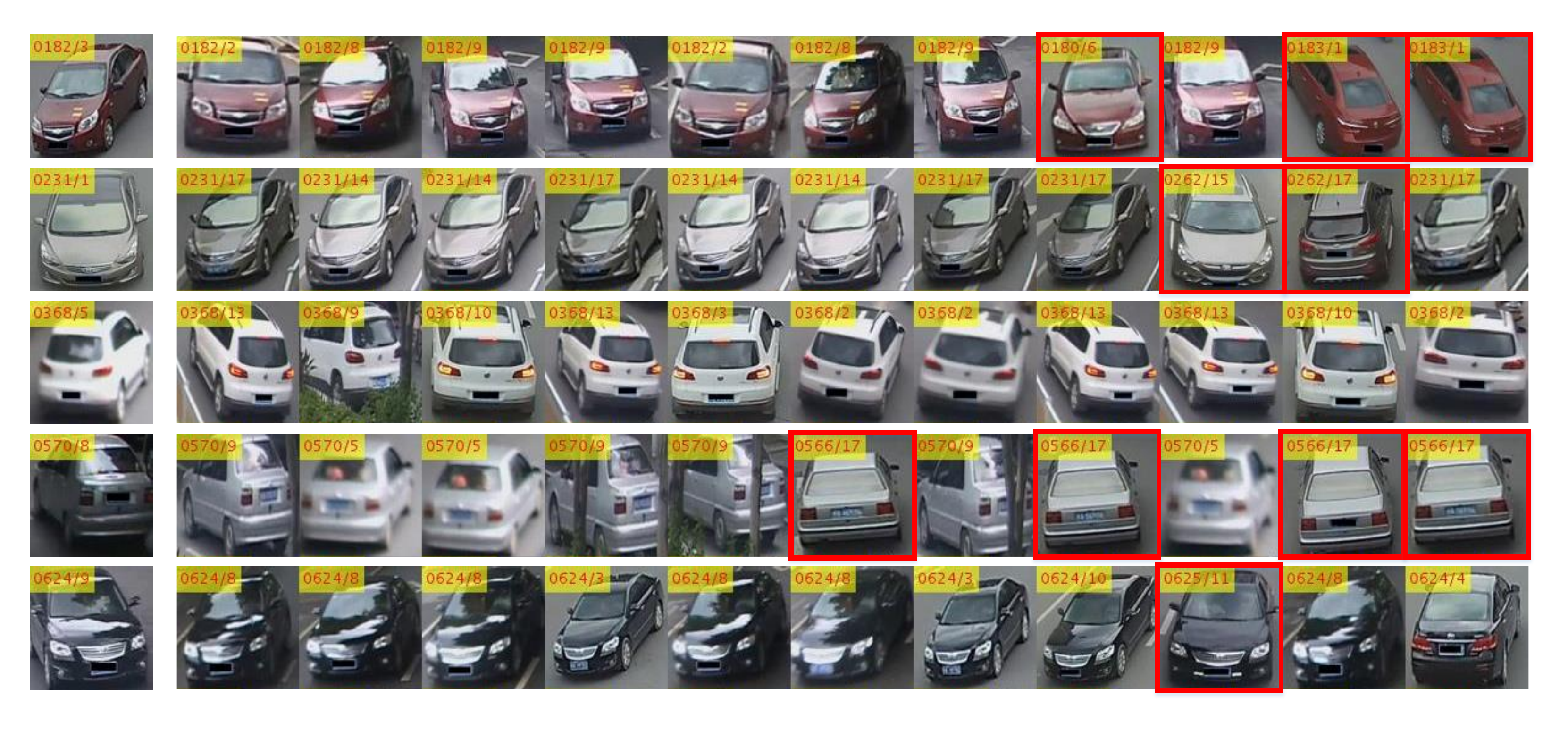}
	\caption{The retrieval results on the VehicleID. The left column shows query images while the images of right-hand side are retrieval results obtained by proposed method} \label{fig10}
\end{figure}

Furthermore, to illustrate the validate of the proposed method, some experiment results on VeRi-776 are visualized. Examples are shown in Fig.\ref{fig10}. In Fig.\ref{fig10}, The left column shows query images, while images on the right-hand side are the top-11 results obtained by the proposed method. Vehicle images with red border are wrong results while other images are right results. For all results, the number on the left-top means Vehicle ID\ Camera ID. The same Vehicle ID represents the same vehicle. The Camera ID is the camera number that images are captured. From Fig.\ref{fig10}, it is significant that our proposed method has high accuracy and good robustness to different viewpoints and illumination.

\section{Conclusion}

In this paper, we propose tiplet center loss based part-aware model for learning discriminative features in local part regions, which increasing discriminative for nuances of vehicles with similar colors and shapes produced by the same or different manufacturers. In addition, an external memory module is adopted to obtain the feature distribution of a local region of each vehicle to perform invariance constraints within the class. The intra-class and inter-class differences are obtained through the tiplet center loss, which improves the performance of vehicle re-identification. Therefore, in our studies, we mainly pay attention to the differences in local details of vehicles, which has an important impact on the task of vehicle re-identification extracted by part fine-grained features.

\section*{Acknowledgements}
This work was supported in part by the National Natural Science Foundation of China Grant 61370142 and Grant 61272368, by the Fundamental Research Funds for the Central Universities Grant 3132016352, by the Fundamental Research of Ministry of Transport of P. R. China Grant 2015329225300, by the Dalian Science and Technology Innovation Fund 2018J12GX037 and Dalian Leading talent Grant, by the Foundation of Liaoning Key Research and Development Program, China Postdoctoral Science Foundation 3620080307.

\bibliographystyle{elsarticle-num}
\bibliography{mybibfile}

\end{document}